%% file: main.tex
\documentclass{article}
    
    \usepackage{iclr2024_conference,times}

    \usepackage{graphicx} 
    \usepackage{color}
    \usepackage{hyperref}
    
    \input{math_commands.tex}

    \graphicspath{figures/}

    \newtheorem{problem}{Problem}
    \newtheorem{question}{Question}
    \newtheorem{conjecture}{Conjecture}
    
    \newcommand{\RR}{\mathbb{R}}
    \newcommand{\EE}{\mathbb{E}}
    \newcommand{\PP}{\mathbb{P}}
    
    \title{\textsc{Learn2extend}: Extending sequences by retaining their statistical properties with mixture models}
    \author{Dimitris Vartziotis$^{1,*}$, George Dasoulas$^{2,*}$ \& Florian Pausinger$^{1,}$\thanks{The authors contributed equally.}\\
    $^1$~TWT Science \& Innovation. Stuttgart, Germany\\
    $^2$~Harvard University. Cambridge, MA, USA}

    \date{\today}
    
\begin{document}
    
    \maketitle

    \begin{abstract}
       This paper addresses the challenge of extending general finite sequences of real numbers within a subinterval of the real line, maintaining their inherent statistical properties by employing machine learning.  Our focus lies on preserving the gap distribution and pair correlation function of these point sets. Leveraging advancements in deep learning applied to point processes, this paper explores the use of an auto-regressive \textit{Sequence Extension Mixture Model} (SEMM) for extending finite sequences, by estimating directly the conditional density, instead of the intensity function. We perform comparative experiments on multiple types of point processes, including Poisson, locally attractive, and locally repelling sequences, and we perform a case study on the prediction of Riemann $\zeta$ function zeroes. The results indicate that the proposed mixture model outperforms traditional neural network architectures in sequence extension with the retention of statistical properties. Given this motivation, we showcase the capabilities of a mixture model to extend sequences, maintaining specific statistical properties, i.e. the gap distribution, and pair correlation indicators.

    \end{abstract}


    \section{Introduction}
    
    \subsection{Problem formulation}
    
    Point processes~\citep{Cox2018, Daley2003-bk} play a central role in modern mathematics as they are often used in mathematical models of real world phenomena~\cite{Babu1996, Snyder1991, Mohler2011}. However, depending on the particular type of point process, it can be expensive to generate instances of point processes, or to extend given point samples~\cite{Beutler1966, Perman1992}.
    The main aim of this paper is to investigate how to extend a given finite sequence of real numbers contained in a subinterval of the real line so that basic statistical properties of this point set are maintained.
    
    \begin{problem}
    Given a sequence of $N$ points in a subset $[0,T]$ of the real line, extend this sequence such that the pair correlation function as well as the gap distributions are maintained.
    \end{problem}
    
    If we view the input as an instance of a random process, then this is an \emph{extremely} hard statistical problem, i.e., we try to infer an underlying model from a single sample. The main point of our work is to leverage deep learning to approximate a sequence extension model.

    \subsection{Motivation and previous work}

    Modeling point processes has been a long-standing problem, that is ubiquitous in multiple real-world domains, ranging from physics~\cite{Fox1978}, neuroscience~\cite{Johnson1996}, and epidemiology~\cite{Gatrell1996} to telecommunication networks~\cite{tele}, and social networks~\cite{Wasserman1980}. Moving beyond standard statistics practices~\cite{Stoyan2006}, machine learning advancements have started to be more imminent into the realm of modeling of point processes~\cite{NEURIPS2018_5d50d227,pmlr-v70-urschel17a,neural_hawkes, zuo2020}. A key element of multiple machine learning approaches is the estimation of the conditional intensity function (e.g. Poisson processes~\cite{Du2016, omi}, and Hawkes processes~\cite{zuo2020}). Although estimating intensity function has proven to be effective, and efficient in numerous cases (i.e. when the likelihood estimation is tractable), in this work, and following the advancements of~\citet{ifltpp}, we estimate directly the conditional density using mixture models, targeting \textit{flexibility} to \textit{multiple types of sequences}.
    
    Going one step further than the sequence extension with retention of statistical properties, we are interested on how such a model can perform in a  multiple next-terms prediction of well-studied sequences. Motivation for this question comes from recent advances on the prediction of the zeros of the Riemann $\zeta$ function on the critical line; see \cite{vartziotis2018contributions, Kampe, Chen2021, Shanker_2012, Shanker2019, shanker2022}. The authors in~\cite{Kampe} developed a machine learning model based on neural networks to predict the next zero of the Riemann zeta function, based on a set of previous zeros. This work has two interesting aspects. First, the authors mostly used distances between neighboring zeros as input features in their models. Second, this work exploits the existence of the Riemann Siegel $\zeta$ function, which is, in a way, an underlying real-valued function that governs the location of the zeros.
    Hence, given the Universal Approximation Theorem for neural networks it is not too surprising that this approach worked reasonably well. The network basically learns the underlying function and bases its predictions on that.
    However, what can be done if such an underlying function does not exist or is not known?
    
    The pair correlation conjecture of Montgomery \cite{mont,odlyzko2,odlyzko1, vartziotis2018contributions} predicts a very particular pair correlation structure of the sequence of zeta zeros. Hence, the neural network to predict the location of zeros, is therefore also able to extend a given sequence so that its pair correlation strucutre is maintained.
    Motivated by the fact that this neural network bases its predictions on information about distances of consecutive points, we were led to wonder whether we can train a machine learning model that can extend an arbitrary sequence so that its pair correlation structure is maintained - however, without the knowledge or use of an underlying function. In other words, whether using only one of the two ingredients used for the prediction of zeta zeros is sufficient to extend arbitrary sequences so that statistical properties are maintained.
    This question naturally requires a different approach as we can not rely on evaluations of an underlying real-valued, continuous function governing the locations of the points as input parameters.
    

    \subsection{Results}
    In this study, we explore the application of mixture models to predict the subsequent terms of sequences while preserving certain statistical indicators. Specifically, our emphasis lies on preserving the gap distribution and the pair correlation function of the extended sequences. To demonstrate the model's generalizability, we evaluated its performance on various classes of sequences: Poisson sequences, locally attractive sequences, and the eigenvalues of the circular unit ensemble. 
    
    Our results indicate that the mixture model adeptly extends these sequences, effectively conserving their inherent statistical properties. In comparison to basic neural network architectures designed for multi-step next-term prediction, our auto-regressive mixture model, which samples future sequence terms in batches, exhibits superior performance both in the prediction of terms and in the retention of gap distribution and pair correlation function. Furthermore, we assess the model's inference capabilities through a case study focused on extending the zeroes of the \(\zeta\) function. Our findings reveal that the predicted values closely align with the actual values, with no evident error propagation.

    \subsection{Outlook}
    In Section \ref{sec:prelim} we introduce the statistical descriptors we use as well as the different types of point processes we are interested in. Section \ref{sec:rej} contains a first approach to the extension of point processes, that is the rejection sampling,  which illustrates the main challenges we face. We address these challenges in Section \ref{sec:learning} suggesting an ML-based mixture model, that estimates the conditional density Section~\ref{sec:exp} contains experimental results showing the strength of the proposed models in the retention of the statistical properties of extended sequences. Finally, we conclude this paper in Section \ref{sec:conclusion}.

    \section{Preliminaries} \label{sec:prelim}

    \subsection{Point processes}
    To define point processes, as well as their statistical indicators, we follow~\cite{girotti}. Given an arbitrary collection, $X=(x_n)_{n\geq1} \subset \RR^+$, of points on the real line, we say that a configuration $\mathcal{X}$ is a subset of $\RR$ that locally contains a finite number of points, i.e. $\# (\mathcal{X} \cap [a,b]) < \infty$ for every bounded interval $[a,b] \subset \RR$.
    A locally finite \emph{point process} $\PP$ on $\RR$ is a probability measure on the space of all configurations of points $\{ \mathcal{X}\}$.
    If $P(x_1, \ldots, x_n)$ is a probability function on $\RR^n$ with respect to the Lebesgue measure which is invariant under permutations of the variables, then $P$ defines a point process. The mapping 
    $$A \mapsto \EE[ \#(\mathcal{X} \cap A)],$$
    which assigns to a Borel set $A$ the expected number of points in $A$, is a measure on $\RR$. Assuming there exists a density $\rho_1$ wrt the Lebesgue measure, i.e., the 1-point correlation function, we have
    $$\EE[ \#(\mathcal{X} \cap A)]=\int_A \rho_1(x) dx,$$
    in which $\rho_1 dx$ represents the probability to have a point in the infinitesimal interval $[x, x+dx]$. Furthermore, the two-point correlation function $\rho_2$ (if it exists) is a function of 2 variables such that for distinct points
    $$\rho_2(x_1, x_2) dx_1 dx_2$$
    is the probability to have a point in each infinitesimal interval $[x_i, x_i+dx_i]$, $i=1,2$.
    
    \subsection{Poissonian pair correlations}
    For a Poisson process the number of points in a given set has a Poisson distribution. Moreover, the number of points in disjoint sets are stochastically independent. A homogeneous Poisson process of rate $\mu >0$ is a Poisson process on $\RR^+$ with intensity measure $\mu \lambda$ in which $\lambda$ is the Lebesgue measure on $\RR^+$. In our one-dimensional setting, the rate of the Poisson process can be interpreted as the expected number of points in a unit interval.
    Hence, the number of points in any interval of length $t$ is a Poisson random variable with parameter $\mu t$.
    There is an important characterisation of homogeneous Poisson point processes in terms of its inter-point distances (or gaps), $x_n - x_{n-1}$, with $x_0=0$.
    The so-called Interval Theorem \cite[Theorem 7.2]{LastPenrose} shows that a point process on $\RR^+$ is Poissonian with rate $\mu >0$  if and only if its gaps $x_n - x_{n-1}$, for $n\geq 1$ are independent and exponentially distributed with parameter $\mu$. Moreover, it is shown \cite[Eample 8.10]{LastPenrose} that the two-point correlation function of a stationary Poisson process is given by the constant function $\rho_2 = 1$.
    
    \subsection{Determinantal point processes}
    We follow \cite{girotti}. A point process with correlation functions $\rho_k$ is determinantal if there exists a kernel $K(x,y)$ such that for every $k$ and every $x_1, \ldots, x_k$ we have
    $$\rho_k(x_1, \ldots, x_k)= \det [K(x_i, x_j)]_{i,j=1}^k.$$
    The kernel $K$ is called correlation kernel of the determinantal point process. The particular structure of a determinantal point process allows to compute the gap probabilities. We refer to \cite[Section 3]{girotti} for the general process and to the next subsection for a particular example.

    \subsection{Circular unitary ensemble (CUE)}
    In 1963 Dyson derived the two-point correlation function for the eigenvalues of unitary matrices. Therefore, let $A$ be an $N \times N$ unitary matrix; i.e. $A \in U(N)$. Denote the eigenvalues of $A$ by $\exp(i \theta_n)$, where $1\leq n \leq N$ and $\theta_n \in \mathbb{R}$. 
    We write
    \begin{equation} \label{cue} \phi_n = \theta_n \frac{N}{2\pi} \end{equation}
    for the normalized (unfolded) eigenphases (Note that $\theta_n$ is considered modulo $2\pi$, but not $\phi_n$!).
    We define
    $$ F(\alpha, \beta,A,N) = \frac{1}{N} \# \{ \phi_n, \phi_m \in [0,N] : \alpha \leq \phi_n - \phi_m < \beta\}.   $$
    We use the Haar measure, denoted as $dA$, (the natural invariant measure) on $U(N)$ to define $$ F_{U}(\alpha, \beta,N) =\int_{U(N)}  F(\alpha, \beta, A, N) dA,$$ in which $A$ is taken uniformly wrt the Haar measure. Then Dyson proved \cite{dyson} that the limit distribution $$F_{U}(\alpha, \beta) = \lim_{N\rightarrow \infty} F_{U}(\alpha, \beta, N)$$ exists and takes the form $$F_{U}(\alpha, \beta) = \int_{\alpha}^{\beta} (R_{2,U}(x) + \delta(x)) dx$$ where $\delta(x)$ is Dirac's $\delta$-function and
    $$R_{2,U}(x) = 1-\left( \frac{\sin(\pi x)}{\pi x} \right)^2.$$

    \subsection{Empirical gap distributions and pair correlations}
    We give an empirical introduction to gap distributions and pair correlations which is closely related to our algorithmic approach. 
    Let $X=(x_n)_{n\geq1} \subset \RR^+$ be an infinite sequence of points on the positive real line which is normalised so that $\EE(\#(X \cap T)):=\EE(\#(X \cap [0,T]))=T$. 
    Let
    $$F(a,b,T)=\frac{1}{T}\# \left \{ x_n, x_m \in [0,T] : a \leq x_n - x_m < b \right \},$$
    which measures correlations between pairs of points. If there exists a limit distribution, $F(a,b)$, for $T \rightarrow \infty$, i.e., 
    $$F(a,b)= \underset{T \rightarrow \infty}{ \lim } F(a,b,T),$$
    then 
    $$ F(a,b) = \int_a^b \rho(x) dx,$$
    in which $\rho(x)$ is the two-point correlation function -- note that $\rho$ is written as a function of the distance between two points, instead of a function of a pair of points.   
    
    Moreover, we define $\delta_n:= x_{n+1}-x_n$ to be the $n$-th gap between consecutive points of the sequence. It follows from our assumption that the $\delta_n$ have mean 1 and we define
    $$G(a,b,T)=\frac{1}{\#(X\cap T)}\# \left \{ \delta_i : i \leq \#(X \cap T) , a \leq \delta_i \leq b \right \},$$ 
    with 
    $$G(a,b)= \underset{T \rightarrow \infty}{ \lim } G(a,b,T),$$
    and (if the limit distribution exists)
    $$ G(a,b) = \int_a^b g(x) dx.$$

    \subsection{Montgomery's pair correlation conjecture}
    Using the Riemann-Siegel $\zeta$-function and the Riemann-Siegel $\theta$-function allows for the computation of huge amounts of zeros of the Riemann zeta function; see for example~\cite{odlyzko1}. These large scale computations can also enable the effective numerical investigation of various famous conjectures surrounding the zeros. Montgomery \cite{mont} formulated the following conjecture about the pair correlation of the normalised zeros of the Riemann zeta function:

    \begin{conjecture}
    For fixed $0<a<b<\infty$,
    \begin{align*}
        \frac {\# \{ (\gamma, \gamma'): 0<\gamma, \gamma' \leq T, 2\pi a (\log T)^{-1} \leq \gamma - \gamma' \leq 2 \pi b (\log T)^{-1}\} }{T \log T /(2\pi)} 
        &  \sim  \\  \int_a^b \left (1-\left( \frac{\sin \pi u}{\pi u} \right)^2 \right) du & 
    \end{align*}
    
    as $T\rightarrow \infty$.
    \end{conjecture}
    
    Furthermore, the Gaussian Unitary Ensemble (GUE) hypothesis asserts that the zeroes of the Riemann zeta function are distributed (at microscopic and mesoscopic scales) like the eigenvalues of a GUE random matrix, and which generalises the pair correlation conjecture regarding pairs of such zeroes.
    An analogous hypothesis has also been made for other $L$-functions; see \cite{sarnak2,sarnak}.
    Importantly, this also suggests a distribution for the gaps of normalised zeros of the zeta function.
    
    
    \section{Rejection sampling } \label{sec:rej}
    To further illustrate our problem, we first discuss a basic approach to the extension of sequences in the following, and, in particular, explain why this approach does not work in general.
    
\begin{figure}[t]
    \includegraphics[width=0.5\textwidth]{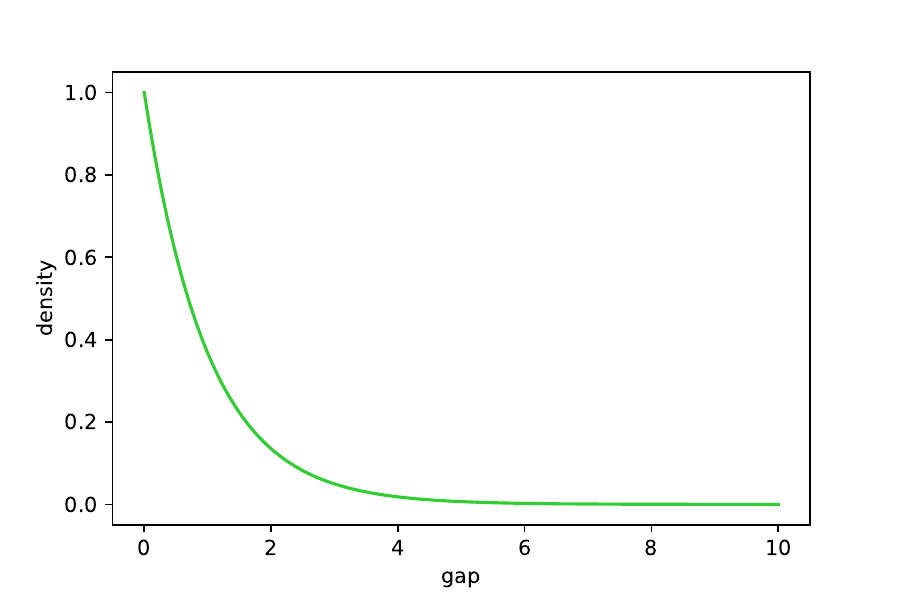}
    \includegraphics[width=0.5\textwidth]{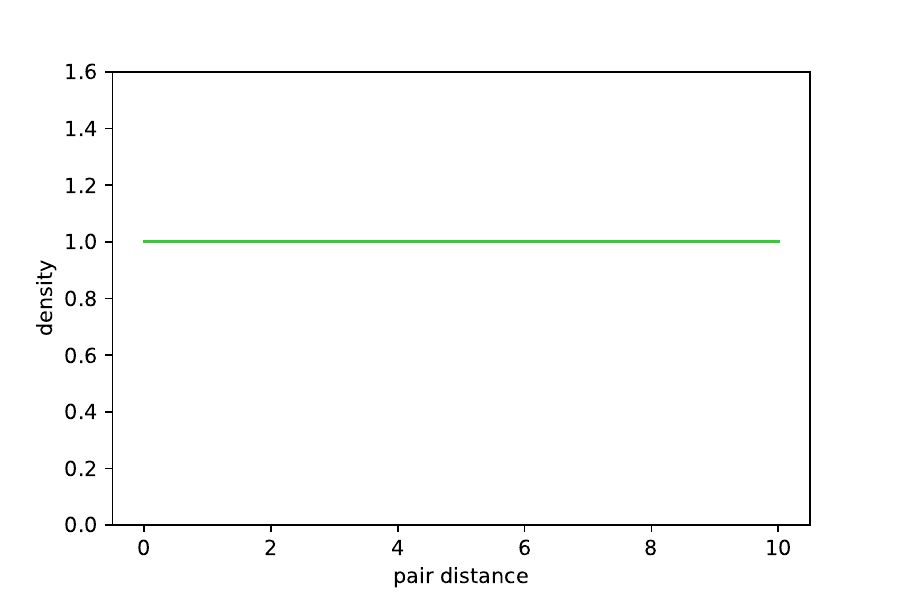}
    \caption{Expected gap distribution function (left) and expected pair correlation function (right) for a Poisson point process. \label{fig:Poisson}}
    \end{figure}

    Importantly, the Poisson point process can be characterised by its gap distribution. 
    The so-called Interval Theorem \cite[Theorem 7.2]{LastPenrose} shows that a point process on $\mathbb{R}^+$ is Poissonian with rate $\mu >0$  if and only if its gaps $x_n - x_{n-1}$, for $n\geq 1$ are independent and exponentially distributed with parameter $\mu$.
    Moreover, it is known \cite[Example 8.10]{LastPenrose} that the two-point correlation function of a stationary Poisson process is given by the constant function $\rho_2 = 1$; see Figure \ref{fig:Poisson}.
    
    While the Poisson point process is characterised via its gap distribution function, a similar result does not exist for eigenvalues of the CUE or GUE. 
    In fact, given that the sequences of eigenvalues can be described as a determinantal point process, it is possible to derive the two-point correlation function of these processes and in turn to find the gap distributions. 
    However, there are no theorems that establish the other direction, i.e., start from a gap distribution and derive the two-point correlation function.
    
    Based on this lack of theoretical results, our first experimental question is whether in our concrete practical situations gap distributions can be utilised to extend sequences, i.e., 
    
    \begin{question}
    Are gap distributions sufficient to extend a given point sample such that its pair correlation function is maintained?
    \end{question}
    
    In other words, can we find a, possibly heuristic, algorithm based on the empirical gap distributions of an input sequence to build an extension that preserves the pair correlation function of the input?
    
    \subsection{Empirical Study}
    We use the finite input to determine a discrete approximation to the gap distribution of the sequence and use this approximation  (rescaled to a PDF) in the form of a histogram \textit{target function} to extend our point-set using classical rejection sampling.
    
    To be more formal, given a set $\mathcal{P}$ of $n$ points in $[0,T]$, i.e., 
    $$\mathcal{P}=\{ x_1, x_2, \ldots, x_n\},$$
    such that $x_i<x_j$ for all $i<j$ and $x_i \in [0,T]$ for all $i$, we seek to extend this sequence such that the resulting sequence
    $$x_{n+1}, \ldots, x_{n+k}$$
    has the same (up to a small pointwise error) pair correlation function as the initial sequence.
    Our algorithm can be described as follows:
    \begin{enumerate}
    \item Compute the empirical gap distribution and pair correlations for the sorted input sequence.
    \item Use a histogram approximation of the empirical gap distribution to extend the input sequence via rejection sampling.
    \item Calculate and output the empirical gap distribution and empirical pair correlation histogram of the extended sequence considering new terms only.
    \end{enumerate}
    
\begin{figure}
    \includegraphics[width=0.5\textwidth]{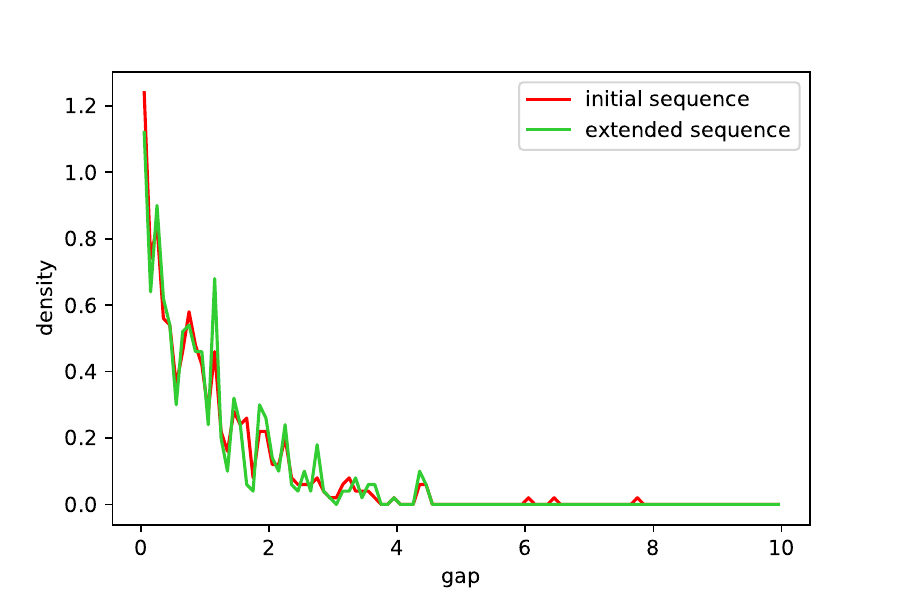}
    \includegraphics[width=0.5\textwidth]{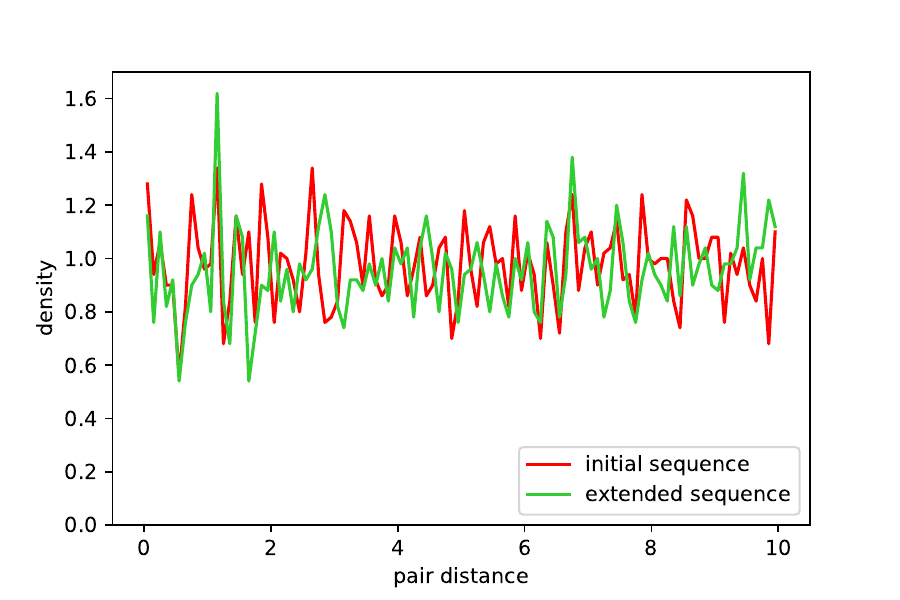}
    \caption{Comparison of empirical gap distribution (left) of one set of 500 Poisson points with its extension (new points only) and of empirical pair correlation function (right) of the same set of 500 Poisson points with its extension.} 
    \label{fig:exp1empirical}
    \end{figure}

    \paragraph{Experiment 1: Poisson Process}
    As a sanity check, we first extend sequences of points coming from a homogeneous Poisson point process. In Figure \ref{fig:exp1empirical} we show a comparison between the empirical gap distribution (left) and the empirical pair correlation function (right) of one input sequence (red) and its extension (green). Overall, and not surprisingly, the algorithm is successful at extending a Poisson sequence to maintaining the gap distribution and pair correlation structure.
    
\begin{figure}[ht]
    \includegraphics[width=0.5\textwidth]{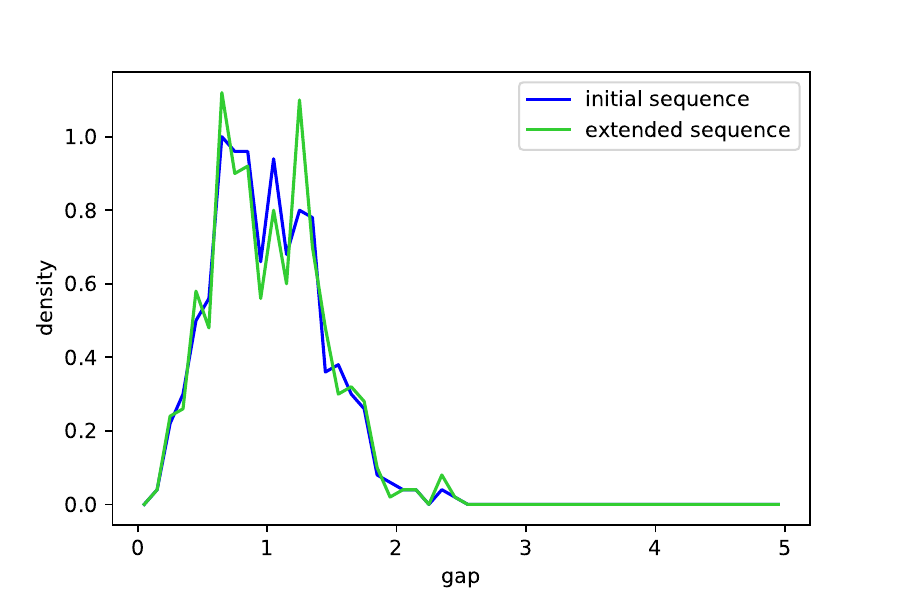}
    \includegraphics[width=0.5\textwidth]{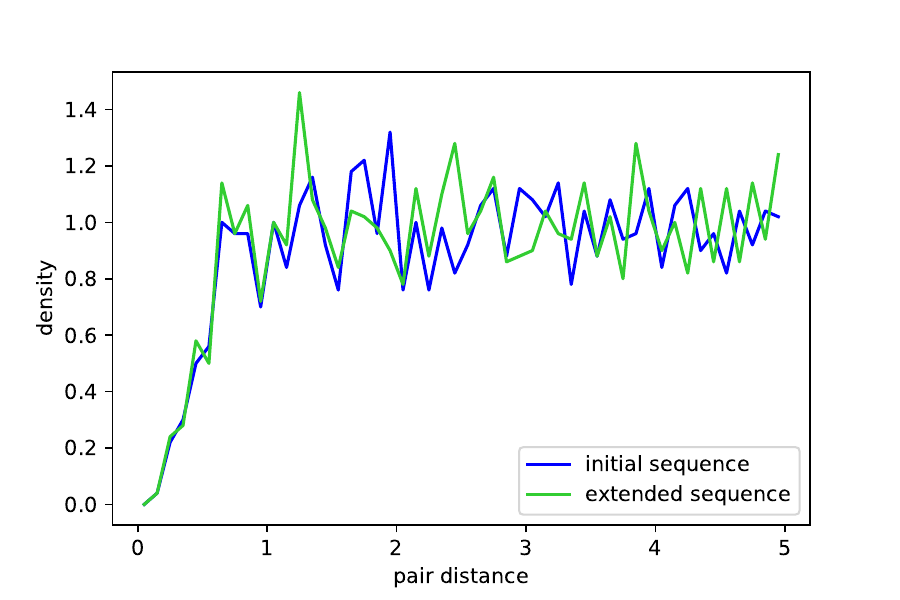}
    \caption{Comparison of empirical gap distribution (left) of one set of 500 CUE points with its extension and of empirical pair correlation function (right) of the same 500 CUE points with its extension. \label{fig:exp2empirical}}
    \end{figure}

\begin{figure}[ht]
    \includegraphics[width=0.5\textwidth]{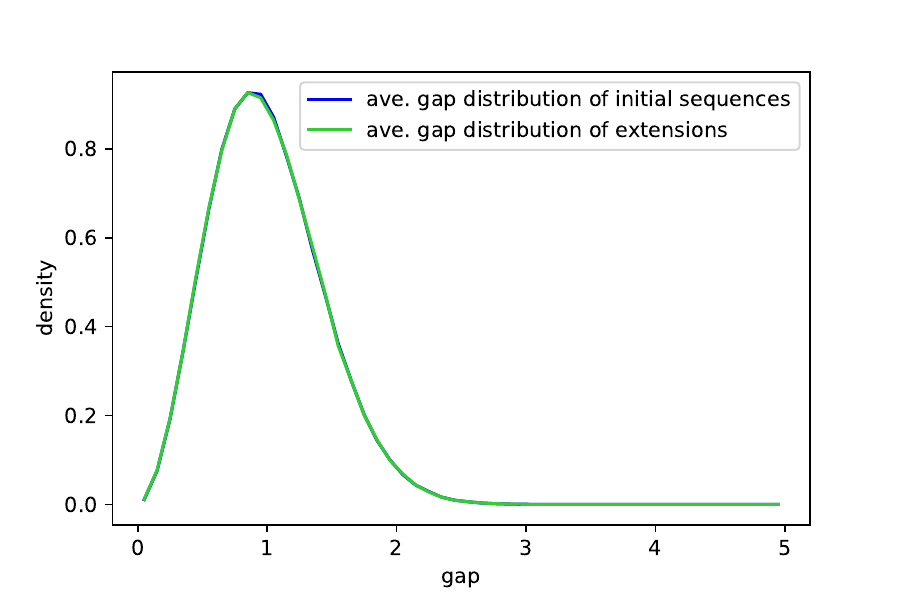}
    \includegraphics[width=0.5\textwidth]{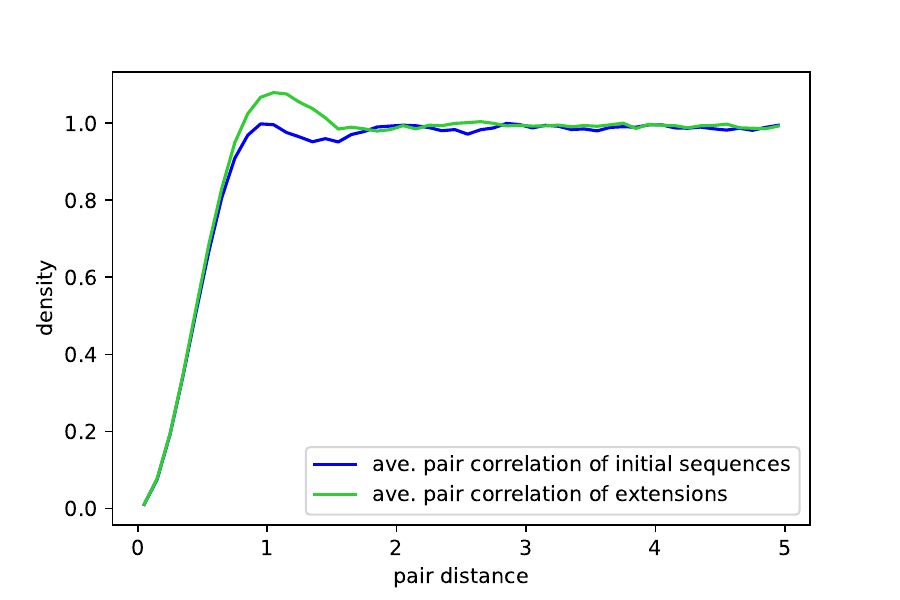}
    \caption{Average gap distributions (left) of 1000 extended sequences and average pair correlation graph (right) of 1000 extended sequences. \label{fig:exp2average}}
    \end{figure}

    \paragraph{Experiment 2: Extension of sequence of eigenvalues of CUE}
    In our second experiment, we extend sequences of normalised eigenphases (see equation \eqref{cue}) of $N \times N$ unitary matrices from the CUE ensemble.
    Figure \ref{fig:exp2empirical} shows a comparison between the empirical gap distribution (left) and the empirical pair correlation function (right) of a single input sequence (blue) and its extension from rejection sampling (green). In Figure \ref{fig:exp2average} we compare the average gap distribution and average pair correlation function of 1000 such sequences from the CUE ensemble with those of their extensions (new points only considered). Again, we see the algorithm is successful at replicating the gap distribution of input sequences in their extensions. 
    
    However, in Figure \ref{fig:exp2average} (right) we see that the average pair correlation of the extended sequences is certainly not a perfect match, with a clear bump present around $x=1$. Hence, this clearly shows that the gap distribution is in general not sufficient to determine the exact pair correlation.
    
    \subsection{Discussion}
    In Experiment 2 we observe a bump at around 1 in the graph of the pair correlation function of the extended sequences that is not present in the input. The immediate conclusion is that we seem to have the approximately right list of gaps, but their order is not correct. In other words, it seems that our local-to-global extension does not work in the case of eigenvalues of the CUE. 
    
    To explore this phenomenon further, we look at the 2-gaps of the input and extension. Note that  it can be shown \emph{that the distribution of the sum of two independent random variables looks like the convolution of the distribution functions.}
    
\begin{figure}[t]
    \includegraphics[width=0.325\textwidth]{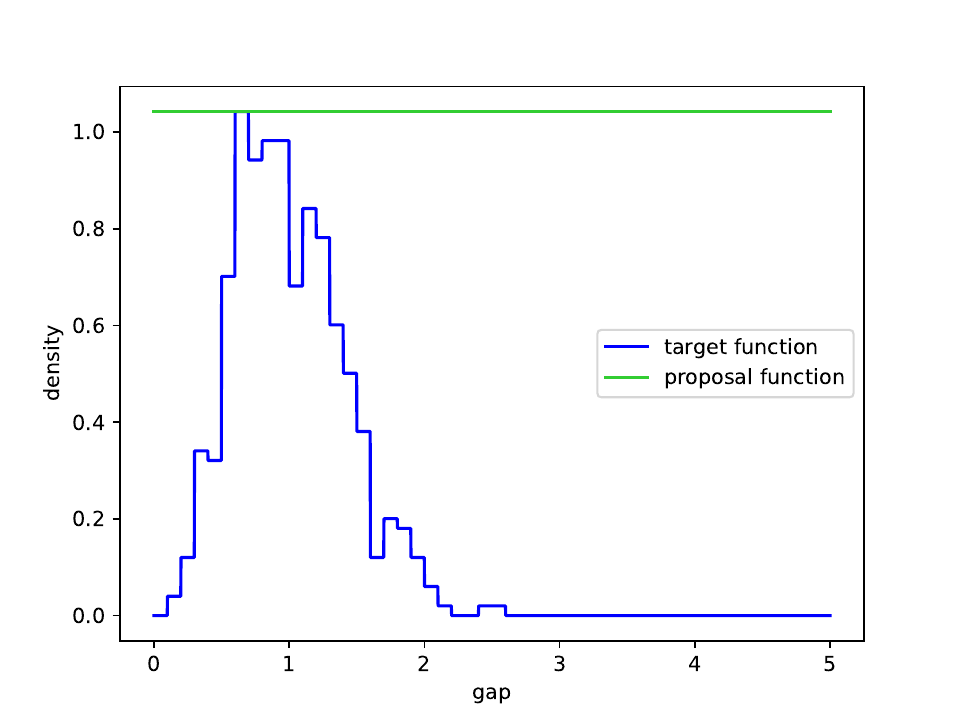}
    \includegraphics[width=0.325\textwidth]{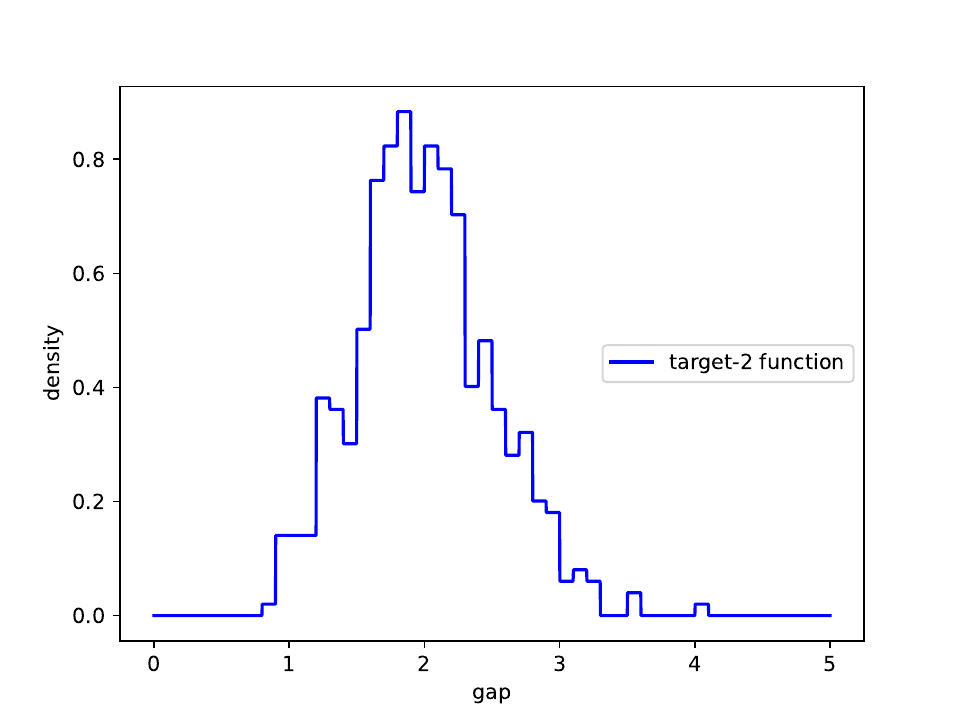}
    \includegraphics[width=0.325\textwidth]{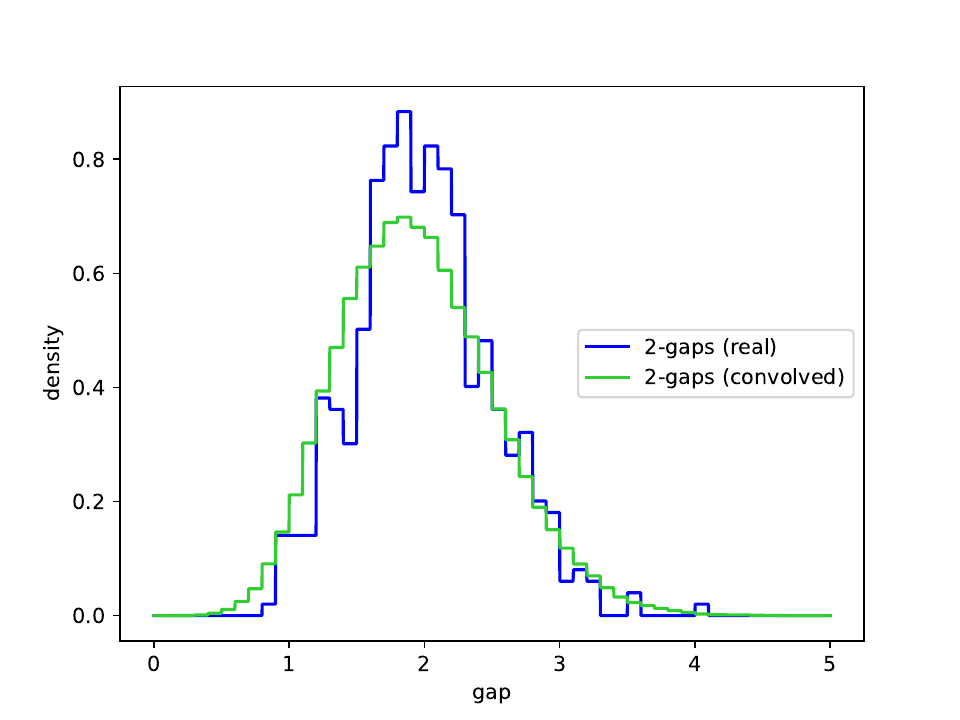}
    \caption{Empirical gap-distribution function of input (left), empirical 2-gap distribution of input (middle), comparison of empirical 2-gap to convolution of gaps (right). \label{fig:twoGapInput} }
    \end{figure}

\begin{figure}
    \includegraphics[width=0.5\textwidth]{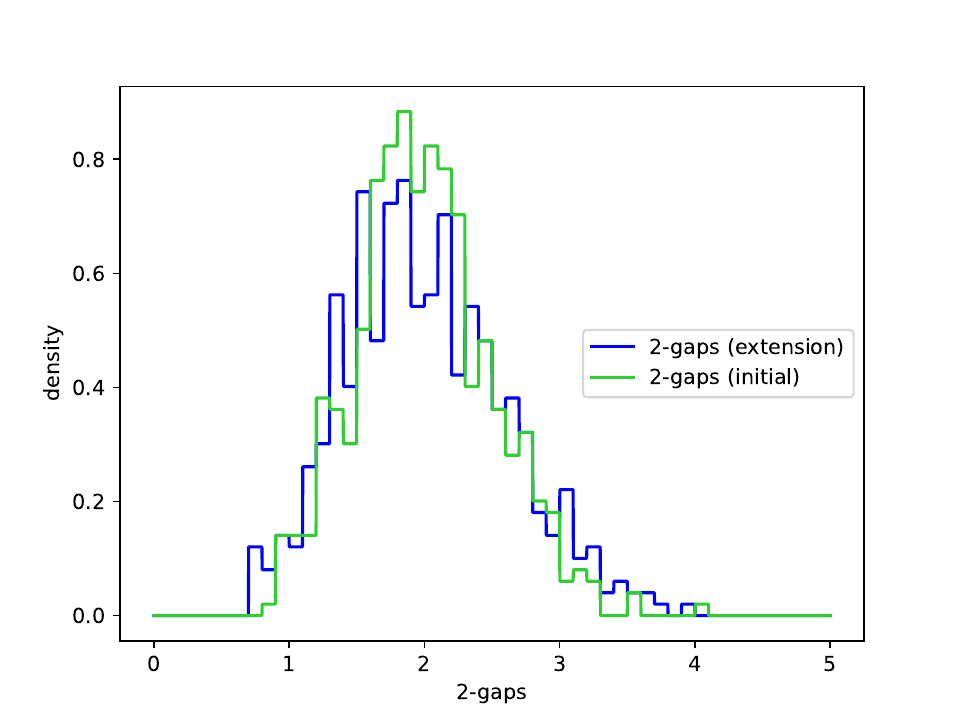}
    \includegraphics[width=0.5\textwidth]{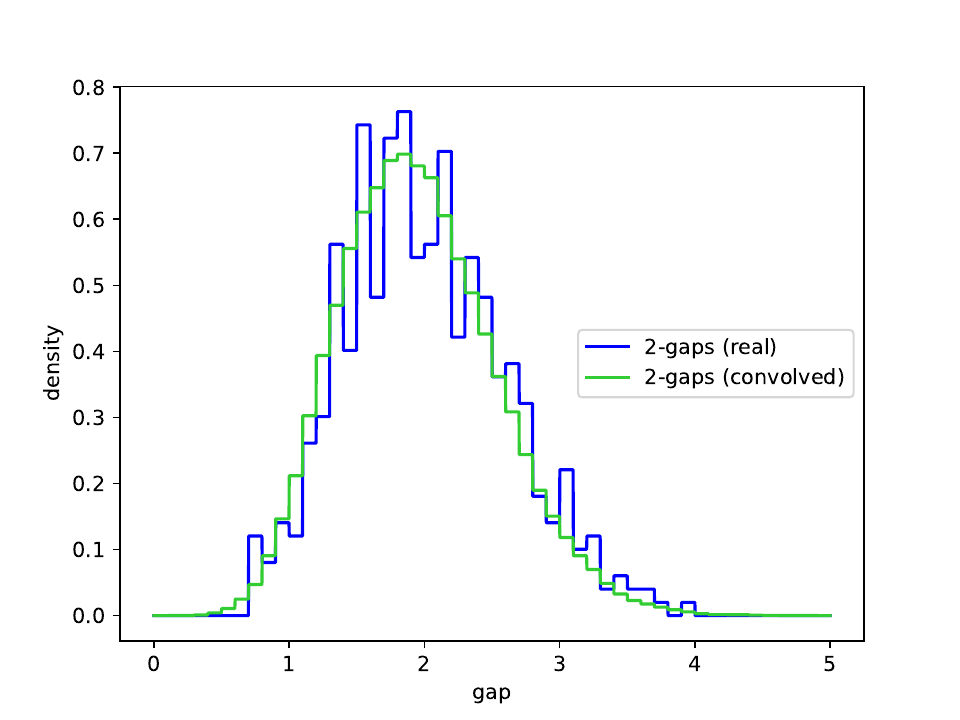}
    \caption{Comparison of empirical 2-gap-distribution function of a single input and extension (left), comparison of empirical 2-gap distribution of a single extension and covolution of input gaps (right). \label{fig:twoGapExtension}}
    \end{figure}

    We observe that our input sequences have a 2-gap distribution {\bf different} from the convolution of its gap distribution with itself; see Figure \ref{fig:twoGapInput}. However, we also observe that the extensions have a 2-gap distribution {\bf similar} to the convolution of its gap distribution; see Figure \ref{fig:twoGapExtension}.
    Hence, the problem with using our extension procedure in Experiment 2 is the assumption that gaps are {\bf independent} while, in reality, there is a dependence corresponding to the local repulsion behavior of the underlying point process.
    
    \section{Learning point processes} \label{sec:learning}
    Machine learning models have emerged as versatile tools for understanding and predicting point processes~\cite{kulesza2012, mariet2015, zuo2020}. By employing algorithms that can recognize intricate patterns within large volumes of data, these models can encapsulate the underlying dynamics of sequential events in both time and space. 
    
    For instance, in the area of social media, predicting the times at which users post or engage can be modeled as a point process~\cite{monti2019fake, Ahmad2020}, and ML can help discern the factors that drive user activity peaks or decreases. Another compelling application is in finance, especially high-frequency trading~\cite{Paiva2019}. The timing of trades, often triggered by a myriad of factors including news releases, market sentiments, or algorithmic strategies, can be seen as a point process. Machine learning models can be used to predict the likelihood of trading events within minuscule time frames, offering traders insights into immediate future market activities.
    
    \subsection{Learning with Mixture distributions}\label{seq:mixture_intro}

    Moving beyond clustering applications~\cite{celeux2018model}, we can apply mixture models to estimate the conditional density of point processes, and more particularly of temporal point processes, as noted by~\cite{ifltpp}.
    Consider a general finite sequence $S = (s_1, s_2,...,s_i,...,s_N)$, In this context, modeling the individual terms of the sequence $s_i$ is parallel to modeling the event times $t_i$ in a temporal point process. The primary focus in this analogy shifts to the modeling of the gaps between successive terms in the point process, denoted as $\tau = s_{i+1} - s_{i}$. These gaps, known as inter-event times in the realm of temporal point processes, represent the intervals between consecutive events. 
    Following \citep{ifltpp}'s work, and due to the fact that the sequence gaps of our interest are positive, we employ log-normal distributions as mixture components. For a $K$ number of mixture components, the probability density function of a log-normal mixture can be defined as:
    \begin{align}
        \label{eq:lognorm-mix}
        p(\tau | \rmW, \rmM, \rmSigma) = \sum_{k=1}^K \frac{1}{\sqrt{2\pi}}\frac{\rmW_k }{\tau \rmSigma_k } \exp \left(-\frac{(\log \tau - \rmM_k)^2}{2\rmSigma^2_k} \right),
    \end{align}
    where $\rmW$ are the mixture weights, and $\rmM,\rmSigma$ are the standard statistical measures (i.e. mean, and standard deviation).

Sampling from mixture models typically involves two steps: i) \textit{Component Selection}, where a component from the mixture is sampled using a categorical distribution based on the mixing weights, and ii) \textit{Data Generation}, where once a component is selected, a data point from the chosen component's distribution is sampled. This two-step process can written in closed form as following:
    \begin{align}
        \vz  & \sim \Categorical (\rmW)  \\  \varepsilon  & \sim \Normal(0, 1) \\ \tau &= \exp (\rmSigma^T \vz \cdot \varepsilon + \rmM^T \vz).
    \end{align}
    The samples $\tau$ drawn using the procedure above are differentiable~\citep{jang2016gumbel} through the Gumbel-softmax trick with respect to the means $\rmM$ and scales $\rmSigma$.
    This sampling procedure yields samples that are differentiable, as noted by~\cite{ifltpp} through the application of the Gumbel-softmax technique~\cite{jang2016gumbel}. This differentiability is crucial, particularly with respect to the parameters of interest, $\rmM$ and $\rmSigma$, enabling efficient gradient-based optimization methods.
  
    \subsection{Sequence Extension Mixture Model}
    
    In Section~\ref{seq:mixture_intro}, we have shown how we can sample from mixture models, given the parameters of the distribution $\rmW, \rmM, \rmSigma$. Now, we proceed by describing how we can obtain these parameters through the lens of recursive neural networks, and consequently defining the \textit{Sequence Extension Mixture Model} (SEMM), following the derivations from~\cite{ifltpp}. 
    
    \paragraph{Context vectors} Given a sequence of point process terms $\va = [a_1,..., a_n]$, we define the context vector $\vh_i$ for each time step $i < n$ of the sequence, that is dependent on all the previous sequence terms: $\vh_i = \phi({a_1,...,a_{i-1}}).$ For this modeling, we make use of a recurrent neural network (RNN) that embeds a sequence of terms into a fixed-size vector $\vh_i$. Specifically, we utilize a multi-layer gated recurrent unit (GRU)~\cite{cho2014learning}, that computes the context vectors as following:
    
    \begin{equation}\label{eq:gru}\begin{array}{ll}
        \vr_i = \sigma(\rmR_{1} a_i + \vb_{ir} + \rmR_{2} \vh_{(t-1)} + \vb_{hr}) \\
        \vz_i = \sigma(\rmZ_{1} a_i + \vb_{iz} + \rmZ_{2} \vh_{(t-1)} + \vb_{hz}) \\
        \vn_i = \tanh(\rmN_{1} a_i + \vb_{in} + \vr_i * (\rmN_{2} \vh_{(t-1)}+ \vb_{hn})) \\
        \vh_i = (1 - \vz_i) * \vn_i + \vz_i * \vh_{(i-1)},
    \end{array}\end{equation}
    where $\rmR_1, \rmR_2, \rmZ_1, \rmZ_2, \rmN_1, \rmN_2$ are the weights of the $\vr_i \text{ (reset) },\vz_i \text{ (forget) }, \text{ and } \vn_i \text{ (new) }$ gates respectively.

    \paragraph{Model parameters} After obtaining the context vectors for all observed sequence terms, the parameters of the distribution can be computed as an affine function of $\vh_i.$ Specifically, we have:
    \begin{align}
        \rmM_i &= \Theta^{(1)}_\rmM \vh_i + \Theta^{(2)}_\rmM \\
        \rmSigma_i &= e ^ {\Theta^{(1)}_\rmSigma \vh_i + \Theta^{(2)}_\rmSigma} \\
        \rmW_i &= \softmax(\Theta^{(1)}_\rmW \vh_i + \Theta^{(2)}_\rmW), 
    \end{align}
    where the set of $\rmTheta^{(1)}_{*},\rmTheta^{(2)}_{*},$ for $* \in \{\rmM,\rmSigma,\rmW\}$ are the learnable parameters of the model. Using the learnt distribution parameters, we can now sample from the mixture model, as computed in Equations $(4),(5),(6).$

    \section{Experiments} \label{sec:exp}

    Given the defined mixture model for the extension of point processes, we are proceeding with the evaluation of its performance. The main axis of performance evaluation is the comparison of statistical indicators of the simulated sequences, and the actual ones. Moreover, we showcase a case study on real-valued zeroes of the $\zeta$ function, where we evaluate how well the model can predict multiple next terms in the sequence.
    
    Measuring properly the performance of a sequence extension model requires the utilization of point processes that come from different generators. That is mainly due to the structural similarity of sequences that occurs within one point process class, and the structural difference of sequences that occurs across point process classes. For the experimentation setups of this work, we mainly focus on four types of point processes: a) Circular Unit Ensemble (CUE) eigenvalues, b) Poisson processes, c) locally attractive processes, and d) zeroes of the $\zeta$ function. 
    
    \begin{figure}[h!]
        \centering
        \includegraphics[width = \textwidth]{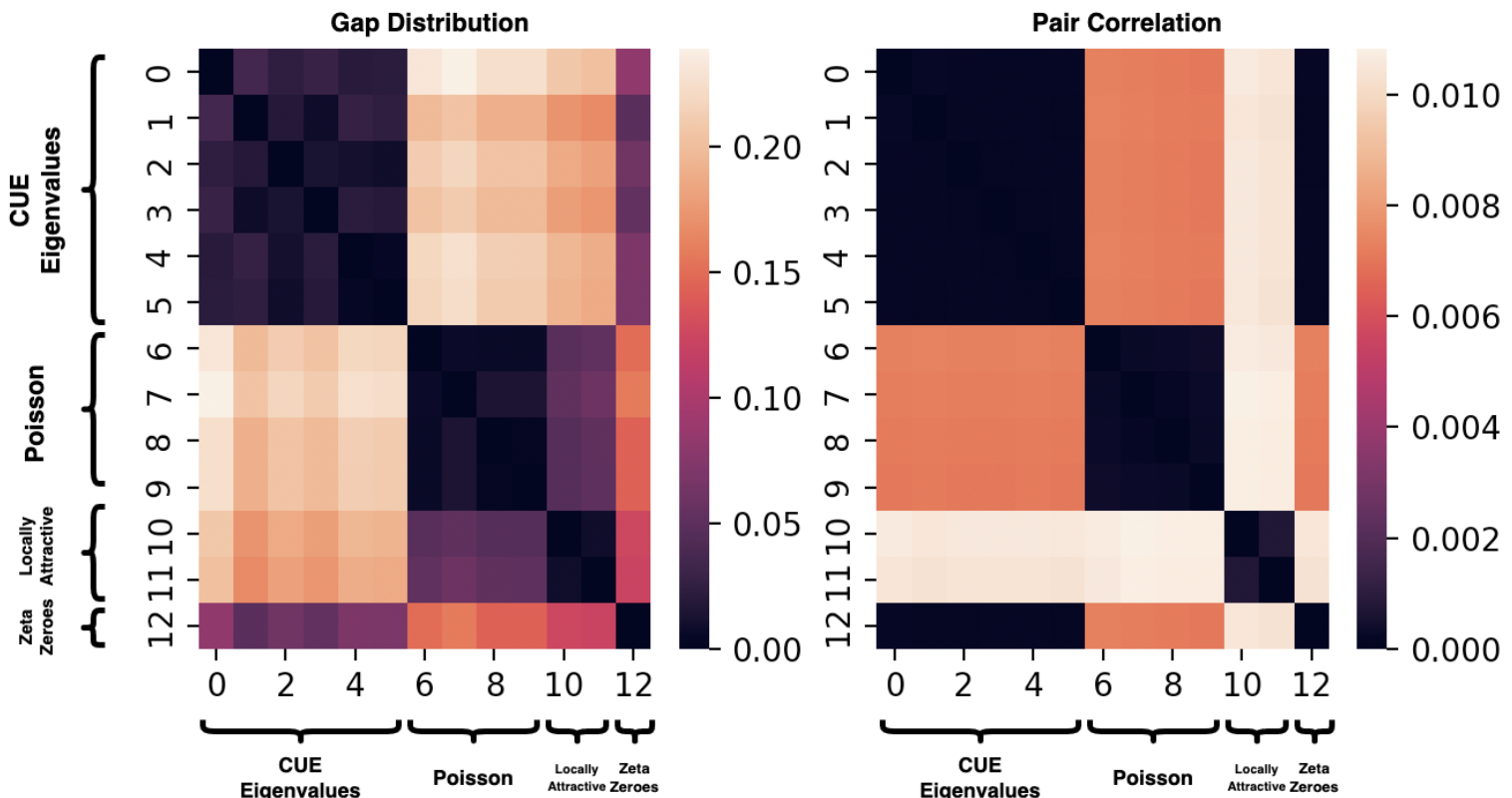}
        \caption{Wasserstein Distance of different point processes.}
        \label{fig:wd}
    \end{figure}
    
    The structural differences of the various types of processes can be quantified through the computation of the Wasserstein distance among all pairs of sequences. On Figure~\ref{fig:wd}, we visualize the Wasserstein distances among the four types of sequences for the statistical indicators: gap distribution and pair correlation function.
    
    \paragraph{Experimental Setup} In the following we briefly describe the different point processes we used to generate our data.
    \begin{itemize}
    \item Poisson process: We sample $N$ points in the interval $[0,500]$ with intensity $\mu=1$.
    \item CUE: We generate an $500 \times 500$ unitary matrix, calculate its eigenvalues, and normalize the sequence of eigenvalues as in (\ref{cue}).
    \item Riemann Zeros: We normalize blocks of $T=500$ consecutive zeros, so that an interval of length $T$ contains approximately $T$ zeros.
    \item Locally attractive points: We generate different Gibbs point processes in the interval $[0,500]$ with density $\mu=3.5$ and custom pair-potential function $h$ with
    $$h(r)=\begin{cases}
        r^4 & \text{ if } r<0.85 \\
        \min \left(0.85^4, \exp(1.4-1.4r)\right) & \text{ else }
    \end{cases}$$
    \end{itemize}
    As illustrated in Figure \ref{fig:wd} sequences of Riemann zeta zeros and sequences of eigenvalues of unitary matrices are structurally very similar. This observation is formalised in the before mentioned pair correlation conjecture of Montgomery and was numerically investigated in \cite{odlyzko1,odlyzko2}. On the other hand, the locally attractive points are very different from the CUE and Riemann points which are known to be locally repulsive. The Poisson points are neither locally repulsive nor attractive and are as such at an intermediate distance from both, the repulsive and the attractive instances of point sequences.

    \subsection{Modeling pair correlation}\label{sec:exp_pr}
    
    \paragraph{Experiment 3: CUE Eigenvalues revisited} We evaluate the impact of the mixture model on the modeling of another set of point processes, and more specifically the eigenvalues of Circular Unit Ensemble. We know that this class of point processes have locally non-attractive behavior.
    
    \begin{figure}[h!]
        \centering
        \includegraphics[width = \textwidth]{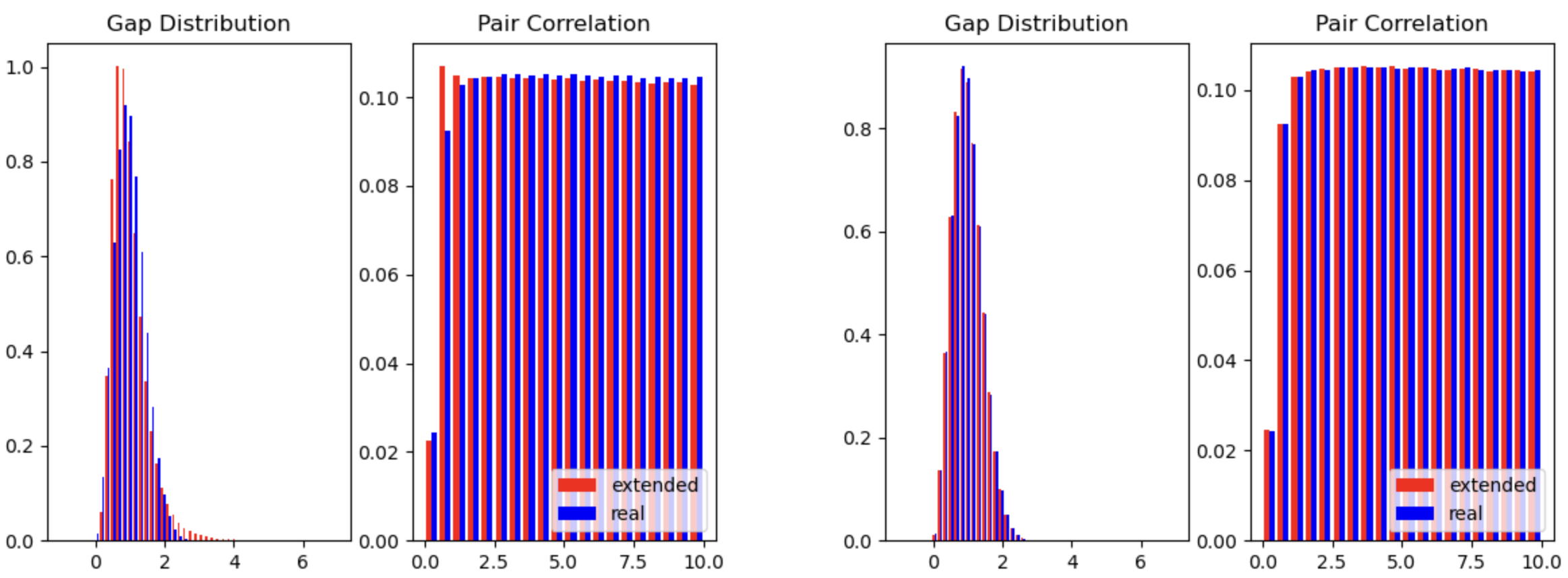}
        \caption{Measured on gap distribution and pair correlation measures for real CUE point processes, and simulated ones. On the left, the simulated point processes are the result of an untrained instance of the model ($\text{epoch = 1}$). On the right, the simulated point processes are given after $200$ epochs of model training.}
        \label{fig:cue}
    \end{figure}
    For this evaluation, we used a set of $500$ sequences derived from CUE computations. Using a standard train/validation/test split, a) we trained SEMM for $200$ epochs, b) we probed the trained model to yield $500$ simulated sequences, and c) we measured the average gap distribution and pair correlation of the simulated sequences. On Figure~\ref{fig:cue} we record the measured statistics in checkpoints: one in the case where the model is not yet trained (the simulated sequences are the result after the 1st training epoch), and  one in the case where the case is fully trained on the train set. As we can observe, the trained model is able to simulate properly the CUE sequences, as both gap distribution and pair correlation are matched. 
    
    \paragraph{Experiment 4: Poisson Process revisited} Next, we repeat the evaluation setup for the class of Poisson processes. Here, we give two examples of processes, specifically the \textit{a) non-stationary}, and \textit{b) stationary} \textit{Poisson processes}. Following the same experimental protocol with CUE sequences, we probe the trained SEMM model to extend the two types of sequences for a set of $500$ next terms.
    
    \begin{figure}[h!]
        \centering
        \includegraphics[width = \textwidth]{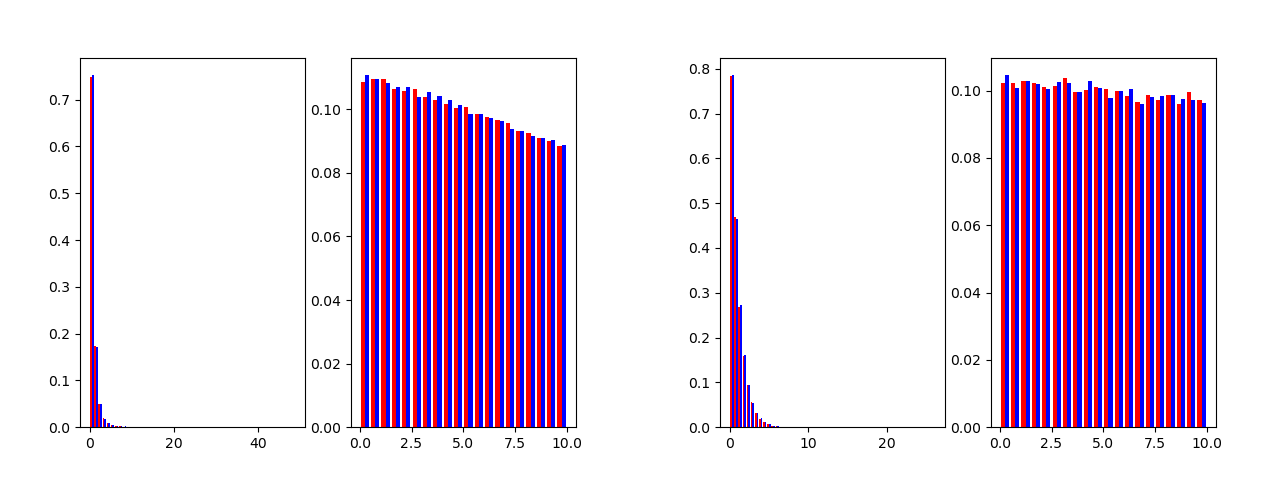}
        \caption{Statistics plots for actual versus predicted values for non-stationary (left) and stationary (right) point processes.}
        \label{fig:poisson_pr}
    \end{figure}
    
    We can observe on Figure~\ref{fig:poisson_pr} that the trained SEMM model is able to model and extend accurately the Poisson processes. Specifically, in the case of the non-stationary processes, SEMM is able to replicate the monotone behavior of the pair correlation function.    
    
    \paragraph{Experiment 5: Locally Attractive Processes} Similarly to the CUE and Poisson processes, we next probe the simulation capabilities of SEMM for locally attractive processes. Once again, we utilize $500$ actual sequences, and we simulate $500$ next terms. 
    \begin{figure}[h!]
        \centering
        \includegraphics[width = .7\textwidth]{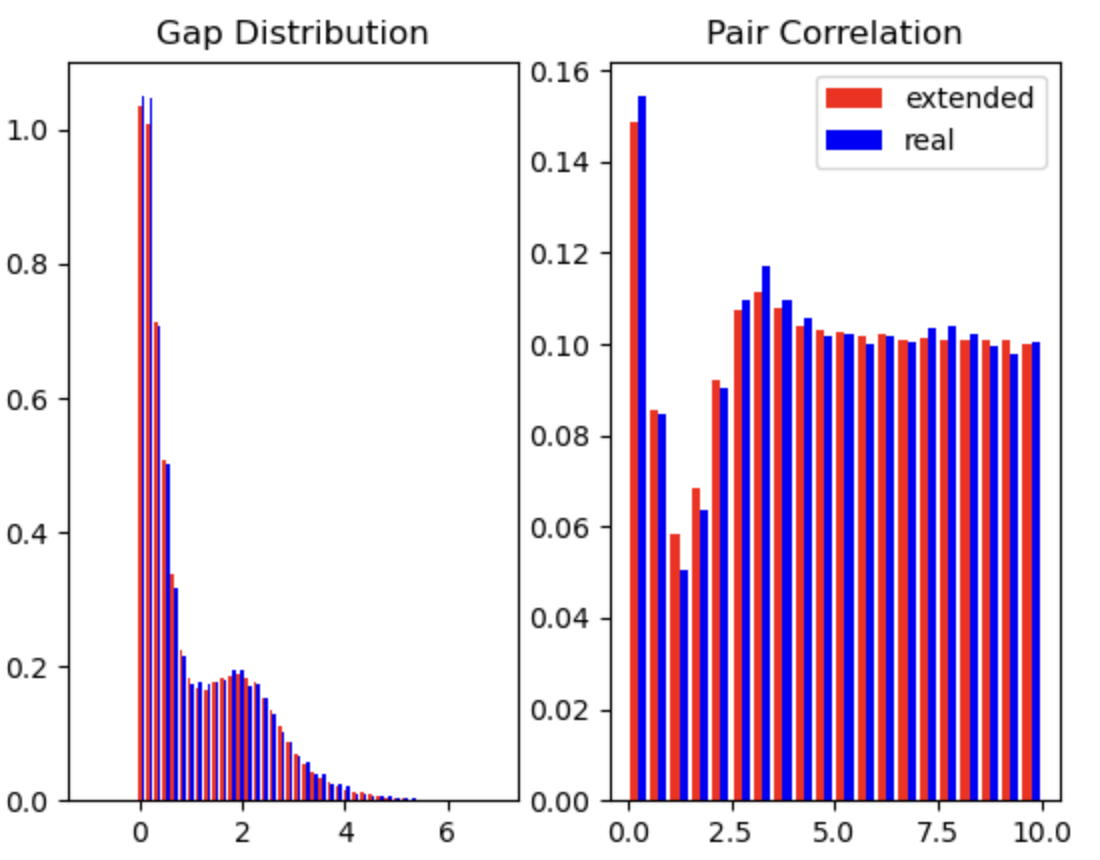}
        \caption{Statistics plots for actual versus predicted values for locally attractive sequences.}
        \label{fig:locally_attractive_pr}
    \end{figure}
    On Figure~\ref{fig:locally_attractive_pr}, we observe the simulated gap distribution and pair correlation. We can note that the model yields slightly worse statistical indicators, with respect to CUE and Poisson processes, showcasing the hardness of modeling locally attractive processes.
    
    \paragraph{Ablation Study} In Table~\ref{tab:ablation}, we show the ablation study of the Sequence Extension Mixture Model with respect to the size of the context representation, as well as the number of mixture components for the case of gap distribution. The first conclusion from the results is that different configurations are suitable for each sequence category. Specifically, while for the CUE, and the locally attractive sequences, the higher number of mixture components ($=64$) was more effective, for the case of the Poisson sequences, $32$ mixture components suffice. This can be related to the fact that the structural complexity of the gap distribution function seems lower than in the case of CUE, and locally attractive sequences.

    \paragraph{Comparison with Neural Networks} In order to show the superiority of the mixture models for modeling sequences, we benchmark a case of neural network architectures that perform multiple next-term prediction (500 multiple steps, following the sampling size of future samples of LMNN). Specifically, we utilize:
    \begin{itemize}
        \item[i)] a fully connected neural network (FCNN) model with 5 layers, that takes as input features the first 500 observed values of the simulated sequences. The output is a multi-label vector (\# labels = 500), that describes the extended sequence terms.
        \item[ii)] a gated recurrent neural network model (GRU), following Equation~\ref{eq:gru}. The basic difference between the GRU architecture, and LMNN model is that the latter learns the parameters of the generating distribution, while the former learns to predict directly the next terms, given the observed values.
    \end{itemize}

    In Table~\ref{tab:ablation}, we incorporate the results for the two architectures for comparison with the LMNN configurations. We can observe that both models fail to reach the simulation performance of the mixture models. Especially, FCNN exhibits the worst behavior, showcasing the inability of multi-step predictors to learn future terms of long horizon.

    \begin{table}[] \label{tab:ablation}
    \centering
        \resizebox{\textwidth}{!}{
    
    \begin{tabular}{c|cc|cc|cc}
    \hline
      & \multicolumn{2}{c|}{\textbf{Poisson}} & \multicolumn{2}{c|}{\textbf{CUE}} & \multicolumn{2}{c}{\textbf{Attractive}} \\ \hline 
      \textbf{Model} & \textbf{RMSE ($\downarrow$)}  & \textbf{$P_R$ ($\uparrow$)}   &  \textbf{RMSE ($\downarrow$)}  & \textbf{$P_R$ ($\uparrow$)} & \textbf{RMSE ($\downarrow$)}  & \textbf{$P_R$ ($\uparrow$)} \\ \hline
     FCNN  & 0.282 $\pm$ 0.031  & 0.59 $\pm$ 0.01 & 0.310 $\pm$ 0.14  & 0.53 $\pm$ 0.04 & 0.298 $\pm$ 0.062  & 0.54 $\pm$ 0.03 \\
     GRU  & 0.238 $\pm$ 0.012 & 0.71 $\pm$ 0.01 & 0.256 $\pm$ 0.050  & 0.66 $\pm$ 0.02  & 0.221 $\pm$ 0.055   & 0.71 $\pm$ 0.02 \\
     LMNN - (64-16) & 0.160 $\pm$ 0.020 & 0.84 $\pm$ 0.02 & 0.185 $\pm$ 0.018  & 0.80 $\pm$ 0.01 & 0.200 $\pm$ 0.021  & 0.77 $\pm$ 0.01 \\
     LMNN - (64-32) &\textbf{ 0.133 $\pm$ 0.015} & \textbf{0.86 $\pm$ 0.02} & \textbf{0.170 $\pm$ 0.016}  & \textbf{0.81 $\pm$ 0.02} & 0.212 $\pm$ 0.023  & 0.76 $\pm$ 0.01 \\ \hline
     LMNN - (16-64) & 0.170 $\pm$ 0.018 & 0.83 $\pm$ 0.01 & 0.195 $\pm$ 0.017  & 0.78 $\pm$ 0.02 & 0.210 $\pm$ 0.020  & 0.76 $\pm$ 0.02 \\ 
     LMNN - (32-64) & 0.165  & 0.84 $\pm$ 0.02 & 0.190 $\pm$ 0.015  & 0.79 $\pm$ 0.02 & 0.205 $\pm$ 0.019  & 0.77 $\pm$ 0.02 \\ 
     LMNN - (64-64) & 0.145 $\pm$ 0.021 & 0.85 $\pm$ 0.01 & 0.194 $\pm$ 0.017  & 0.78 $\pm$ 0.01 & \textbf{0.188 $\pm$ 0.018}  & \textbf{0.79 $\pm$ 0.01}\\ 
    
     \hline
    \end{tabular}}
    \caption{Ablation study of LMNN with respect to size of context vectors, and number of mixture components and comparison with neural network architectures for multi-step prediction. We evaluate the Root Mean Squared Error (RMSE), and the Pearson's Correlation Coefficient for each configuration, and sequence category. The configurations are denoted as: LMNN- (\#context size, \#num. mixture components).} 
    \end{table}

    
    
    
    
    \subsection{Case study: Learning zeroes of the $\zeta$ function}
    
    One particular case where such model predictions can be useful is the modeling of the real-valued zeroes of the $\zeta$ function~\cite{vartziotis2018contributions}. The majority of the ML-based approaches for the extension of the $\zeta$ function zeroes~\cite{Kampe} are based on the next-term prediction, where given a subsequence of $m$ terms, the objective is to predict the $m+1$ term. However, such an objective can become trivial, when the ML models can be overfitted with a large amount of sequence terms for training. Here, we test the performance of LMNN model in a more difficult setup: the multiple-step prediction, where given $m$ sequence terms, we predict the $k$ next terms with $k>>1.$ Similarly to Section~\ref{sec:exp_pr}, we use a set of $500$ cropped, and normalized sequences of the $\zeta$ function zeroes for training. Then, given as input a zeroes subsequence with length $=300$ terms, we probe the model to simulate the next $300$ terms.
    \begin{figure}[h!]
        \centering
        \includegraphics[width = \textwidth]{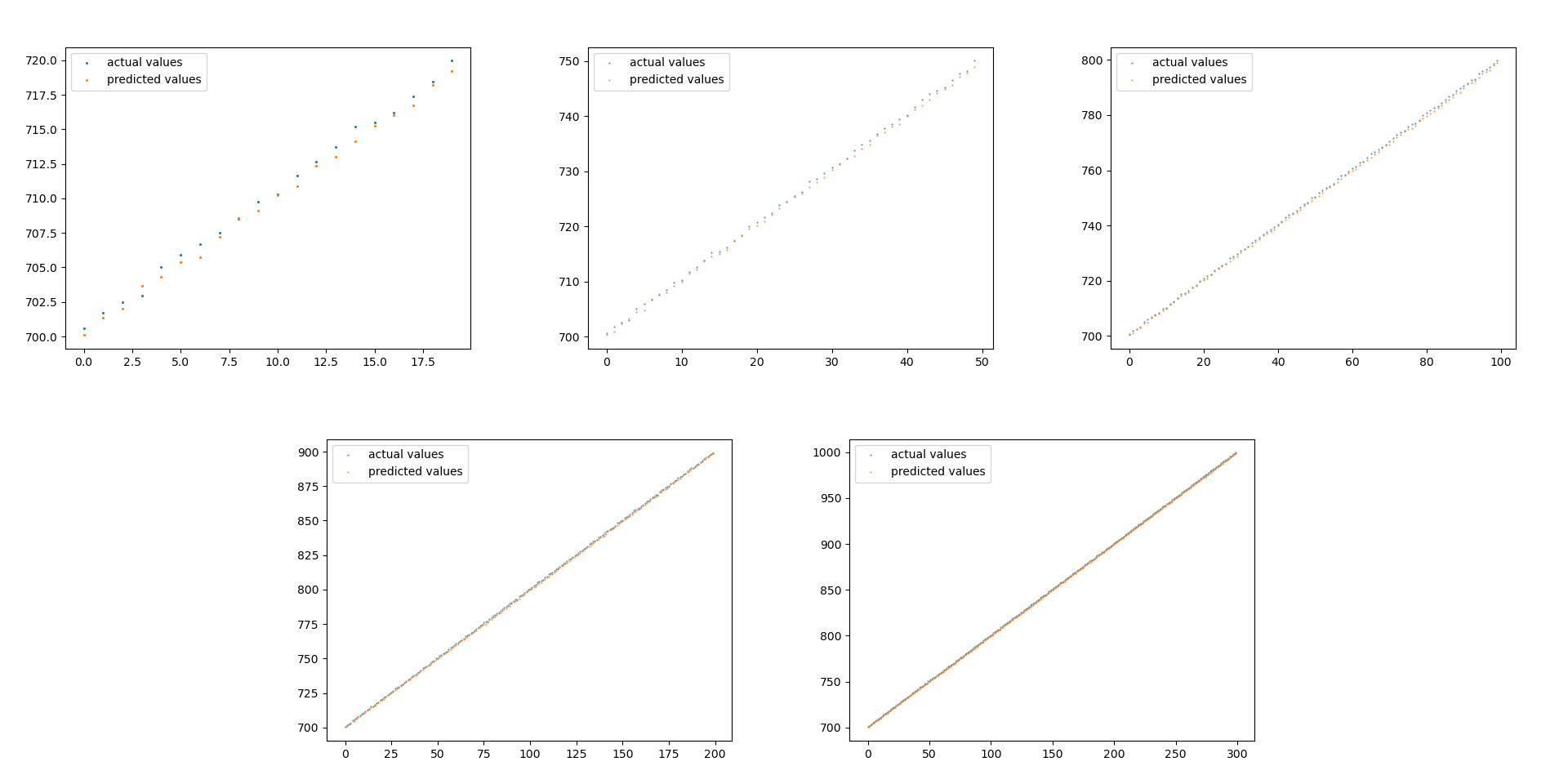}
        \caption{Comparison of actual real-valued zeroes of the $\zeta$ function, and the predicted ones. Starting from left to right, the number of predicted values is increasing, from $20$ up to $300$ next-term values.}
        \label{fig:zeta_zeroes}
    \end{figure}
    
    On Figure~\ref{fig:zeta_zeroes}, we present the comparison plots between the actual real-valued zeroes and the predicted ones for an increasing amount of data points $\in \{20,50,100,200,300\}$. A quite interesting observation is that even for the furthest points of prediction, the estimation error remains close to constant. This highlights a superior advantage of the model, where although the prediction errors should aggregate moving from one term to the next one, the difference in future-step predictions remains independent. 
    
    \section{Conclusion} \label{sec:conclusion}

  In this work, we explore the potential of machine learning for point processes and its contribution to modern mathematics. We build upon a mixture model-based method to extend sequences of real numbers while preserving their statistical properties. Through rigorous experimentation, we evaluated the mixture model's efficacy on a diverse range of sequences, from Poisson sequences to the eigenvalues of the circular unit ensemble. Our findings underscored the model's superior performance in both term prediction and retention of gap distribution and pair correlation function, especially when juxtaposed with traditional neural network architectures. A standout case study on the zeroes of the $\zeta$ function further solidified our model's prowess, showcasing its ability to predict values with remarkable accuracy.

  \section{Acknowledgements}
  We would like to thank Dr. Michael Keckeisen, TWT, for the valuable discussions.
  
    \newpage
    \bibliography{main}
    \bibliographystyle{iclr2024_conference}

    \end{document}

%% file: math_commands.tex
\usepackage{amsmath,amsfonts,bm}

\usepackage{amsmath,amsfonts,amssymb,bm,xspace,nicefrac}

\newcommand{\Categorical}{\operatorname{Categorical}}

\newcommand{\Normal}{\operatorname{Normal}}

\def\eqref#1{equation~\ref{#1}}

\def\1{\bm{1}}

\def\rmM{{\mathbf{M}}}
\def\rmN{{\mathbf{N}}}

\def\rmR{{\mathbf{R}}}

\def\rmW{{\mathbf{W}}}

\def\rmZ{{\mathbf{Z}}}
\def\rmSigma{{\mathbf{\Sigma}}}

\def\rmTheta{{\mathbf{\Theta}}}

\def\va{{\bm{a}}}
\def\vb{{\bm{b}}}

\def\vh{{\bm{h}}}

\def\vn{{\bm{n}}}

\def\vr{{\bm{r}}}

\def\vz{{\bm{z}}}

\DeclareMathAlphabet{\mathsfit}{\encodingdefault}{\sfdefault}{m}{sl}
\SetMathAlphabet{\mathsfit}{bold}{\encodingdefault}{\sfdefault}{bx}{n}

\newcommand{\softmax}{\mathrm{softmax}}

